\documentclass[10pt,twocolumn,letterpaper]{article}

\usepackage[pagenumbers]{wacv} 

\usepackage{epsfig}
\usepackage{graphicx}
\usepackage{amsmath}
\usepackage{amssymb}
\usepackage{booktabs}

\usepackage{diagbox}
\usepackage{multirow}
\usepackage{soul}
\usepackage{color}
\usepackage{array}
\usepackage{ragged2e}

\usepackage{rotating}
\usepackage{makecell}
\usepackage{tabularray}
\usepackage{pifont}
\usepackage[pagebackref,breaklinks,colorlinks]{hyperref}

\usepackage[capitalize]{cleveref}
\crefname{section}{Sec.}{Secs.}
\Crefname{section}{Section}{Sections}
\Crefname{table}{Table}{Tables}
\crefname{table}{Tab.}{Tabs.}

\def\wacvPaperID{1} 
\def\confName{WACV workshop}
\def\confYear{2024}

\begin{document}

\title{QAFE-Net: Quality Assessment of Facial Expressions with Landmark Heatmaps}

\author{ %
    Shuchao Duan\textsuperscript{1}, %
    Amirhossein Dadashzadeh\textsuperscript{1}, %
    Alan Whone\textsuperscript{2}, %
    Majid Mirmehdi\textsuperscript{1}\\
    \textsuperscript{1}School of Computer Science \hspace{0.2cm} \textsuperscript{2}Translational Health Sciences \\
    University of Bristol, UK\\
    {\tt\small \{shuchao.duan, a.dadashzadeh,  alan.whone, m.mirmehdi\}@bristol.ac.uk}
}

\maketitle

\begin{abstract}
{Facial expression recognition (FER) methods have made great inroads in categorising moods and feelings in humans. Beyond FER, pain estimation methods assess levels of intensity in pain expressions, however assessing the quality of all facial expressions is of critical value in health-related applications. In this work, we address the quality of five different facial expressions in patients affected by Parkinson's disease.
We propose a novel landmark-guided approach, QAFE-Net, that combines temporal landmark heatmaps with RGB data to capture small facial muscle movements that are encoded and mapped to severity scores.
The proposed approach is evaluated on a new Parkinson's Disease Facial Expression dataset (PFED5), as well as on the pain estimation benchmark, the UNBC-McMaster Shoulder Pain Expression Archive Database. 
Our comparative experiments demonstrate that the proposed method outperforms SOTA action quality assessment works on PFED5 and achieves lower mean absolute error than the SOTA pain estimation methods on UNBC-McMaster. Our code and the new PFED5 dataset are available at https://github.com/shuchaoduan/QAFE-Net.}
 \end{abstract}

\section{Introduction}
{Certain brain disorders, such as Parkinson's Disease (PD), can affect both voluntary and involuntary muscle movement, resulting in reduced facial expressivity, amongst other issues. Neurologists thus recognise facial expressions as a reliable signal for identifying the severity of symptoms for diagnosis \cite{Williams2002FacialEO,mds-updrs2008} that can assist healthcare professionals in making informed decisions regarding treatment options and monitoring the disease progression in controlled trials.}

{Conventionally, clinical PD neurologists score patient movements using the Movement Disorders Society-Sponsored Revision of Unified PD Rating Scale (MDS-UPDRS) \cite{mds-updrs2008}. An automated system would remove rater bias this measure is notorious for, provide more consistent longitudinal outcomes, and save time for both clinicians and patients, especially if tests can be performed at home \cite{Heidarivincheh2021MultimodalCO}.} 

{Several works have quantified the quality of facial expressions, focussing mainly on the degree of (shoulder) pain estimation \cite{pmlr-v116-xu20a,distanceordering2021,Szczapa_2022_trajec}, which differs from the assessment of natural facial expressions for PD. However, they share the common characteristic of having scores corresponding to the data. In the area of Parkinson's Disease, recent works measure the degree of facial tremor in patients as a {\it detector} of PD~\cite{Abrami2021AutomatedCV,liu2023vision}, 
while others \cite{Hou2021AM2,Jakubowski2021ASO,dynamicfeaturesPD,Moshkova2020FacialEE, CalvoAriza2022ClassicalFA,Huang2023AutoDO,Gomez2023ExploringFE} compare PD patients against controls for loss of various expressions. Only a few \cite{grammatikopoulou2019detecting,moshkova2022assessment} have explored the potential of using facial expressions to predict the severity of PD based on MDS-UPDRS scores.
} 

\begin{figure}
    \centering
    \includegraphics[width=\linewidth]{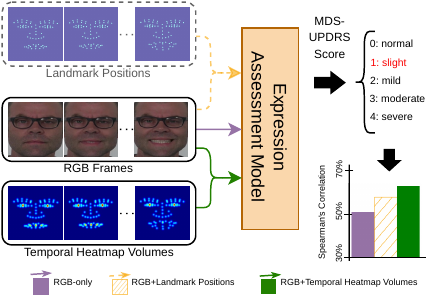}
    \caption{{Our goal is to measure the quality of facial expressions based on clinical scores, i.e. MDS-UPDRS \cite{mds-updrs2008} for assessing severity. To better capture subtle facial muscle movements, we propose temporal facial landmark heatmaps combined with RGB (green), as opposed to just simple landmarks and RGB (orange).} } 
    \vspace*{-4mm}
    \label{fig:page1}
\end{figure}

Facial feature extraction has a long history, in particular based around the use of landmark positions and models that localise specific facial points, on and around the eyes, nose, lips, chin, etc. \cite{Wu2018FacialLD,Zhou_2023_STARLoss}. They offer a pixel-level opportunity to address rigid and non-rigid facial deformations and expressions. While some studies regress precise locations of landmarks from candidate heatmap representations \cite{Wang2022LearningTD,Wan2023PreciseFL}, we only need a reasonably localised 
set of landmarks, from which we generate landmark heatmaps. We hypothesize that such heatmap regions allow us to better encode facial muscle movements for facial expression quality assessment (see \Cref{fig:page1}).

In this paper, we propose a two-stream architecture, QAFE-Net, that encodes  spatiotemporal RGB and landmark heatmap volume features 
to arrive at a quality of performance score for each facial expression.
For the RGB stream, we adopt the backbone of Former-DFER~\cite{formerDFER}, comprising a convolutional spatial transformer and a temporal transformer. For the temporal landmark heatmap stream, which we introduce to capture small and subtle facial muscle movements, we use the SlowOnly~\cite{PoseC3D_heatmap} backbone. Then, the features from the two streams are fused in a cross-fusion decoder to 
{correlate the RGB and landmarks encodings to enable RGB features to focus on local regions containing subtle facial muscle motions guided by landmark features while landmark heatmap features can capture more of the neighbourhood characteristics via RGB features. Finally, the fused features are passed to an MLP to obtain an overall score for an expression.}

{We evaluate our method through comparative experiments on (a)~PFED5, a new Parkinson's disease Facial Expressions Dataset of 5 actions collected from 41 clinically-diagnosed PD patients with varying symptom levels, and (b)~UNBC-McMaster Shoulder Pain Expression Archive Database \cite{pain_dataset_2011}, a public benchmark {\it with only pain expressions} for pain estimation.} 
 
Our key contributions are as follows:
(i) the QAFE-Net model, which includes the encoding of temporal landmark heatmap representations of facial features to assist in facial expression quality measurement, 
(ii) {a {cross-fusion} block for RGB and landmark heatmap region features to capture both global and local expression characteristics}, 
(iii) a new dataset of PD facial expressions with per expression annotations that can be used for classification and measurement, 
and (iv) powered by the above, experiments and ablations to show the proposed method significantly can outperform recent SOTA action quality assessment methods on the PFED5 dataset, as well as achieve better mean absolute error results on the UNBC-McMaster pain measurement dataset. 


\section{Related Works}
\label{sec:literature}
{We briefly review recent literature that is significant to our work, including studies on  facial expression quality measurement, landmarks for face expression analysis, and action quality assessment.}

{
 \textbf{Video-based Facial Expression Assessment --} Many works have attempted to establish the presence of PD through {\it natural} facial expressions alone \cite{Hou2021AM2,Jakubowski2021ASO,dynamicfeaturesPD,Moshkova2020FacialEE, CalvoAriza2022ClassicalFA,Huang2023AutoDO,Gomez2023ExploringFE}. 
For example, Gomez \etal \cite{dynamicfeaturesPD} utilised  RGB frames and  corresponding fine-grained face meshes comprising 468 facial landmarks to model the movement of different facial regions through action units (AUs) to detect PD.
}
Hou \etal \cite{Hou2021AM2} combined geometric and texture features of mouth and eye regions using traditional machine learning approaches to distinguish patients from healthy controls when smiling. 

{Some studies have estimated pain intensity from {\it pain expressions} ~\cite{pmlr-v116-xu20a,distanceordering2021,Szczapa_2022_trajec}. For instance, Ting \etal \cite{distanceordering2021} proposed a distance ordering loss for pain estimation on the self-reported Visual Analog Scale (VAS), by training an ordinal regression model to store the extracted features into a feature pool, and measuring the distances between the input and each pool element. Xu \etal \cite{pmlr-v116-xu20a} incorporated the Prkachin and Solomon Pain Intensity score as an additional supervision signal at the frame level alongside the VAS score as the training labels, and simultaneously predicted AUs to learn frame-level pain features in a multitask learning setup. The statistics of these features were then used as inputs to the sequence-level feature learning phase to estimate pain level.}

Very few have assessed the severity of PD by way of {\it natural} facial expressions~\cite{grammatikopoulou2019detecting, moshkova2022assessment}. {Grammatikopoulou \etal \cite{grammatikopoulou2019detecting} used 8 landmarks from selfies to model the vertical variances of moving features from three facial parts, left mouth, right mouth, and bottom mouth to estimate the PD severity from facial expressions based on MDS-UPDRS scores.} 
Moshkova \etal \cite{moshkova2022assessment} predicted scores by applying traditional machine learning regression models{, such as Support Vector Machine and
Random Forest} to their facial expression feature vector, comprising 33 features of facial activity AUs and 6 emotional expression features. 
We are not aware of any other works that provide quality scores  (on MDS-UPDRS scale or otherwise) for individual or collection of {PD} facial expressions.



 \textbf{Landmarks for Facial Expression Analysis --}  Facial landmark detection has long allowed for the extraction of important facial characteristics, with some recent works of note such as~\cite{guo2021SCRFD,Jin2020PixelinPixelNT,Deng2020RetinaFaceSM}.   
{Investigating landmark positions also features heavily in facial expression analysis 
\cite{ kollias2020exploiting,ngoc2020facial,ryumina2020facial,szczapa2020automatic}.}
For example, 
{Guo \etal \cite{guo2021SCRFD} introduced their Sample and Computation Redistribution strategy for efficient Face Detection (SCRFD) which automatically redistributes more training samples to the shallow stages and optimises the distribution of computation between the backbone, neck, and head of the model, resulting in enhanced computational efficiency while maintaining high precision for both face detection and landmark estimation.} {Szczapa \etal \cite{szczapa2020automatic} adopted Gram matrices to formulate the trajectories of  landmarks to model the dynamics of facial movements on the Riemannian manifold. They split the face into four regions based on landmarks to exploit the relevance of different facial parts for measuring  pain intensity.} 

Previous works have used various numbers of facial landmarks to aid in facial expression analysis, from $8$ points in \cite{grammatikopoulou2019detecting}
to $468$ points in \cite{dynamicfeaturesPD}. {As  SCRFD~\cite{guo2021SCRFD} has the state-of-the-art performance in detecting low-resolution faces and accurately estimating landmarks of even non-frontal faces,} we apply
{their approach} to generate $106$ landmark positions, but exclude facial boundary points and retain $M=73$ landmarks only. 
Then, these landmarks become centres for $M$ heatmap regions per frame in our work, which are assumed to encompass where subtle facial motions occur.

 \textbf{Action Quality Assessment -- } Quantifying the quality of performance of actions is essential in sports and healthcare applications, amongst many others~\cite{CoRe,2020USDL,aqa_tpt,finediving_tsa,Dadashzadeh2020ExploringMB,PCLN_AQA}. 
Tang \etal \cite{2020USDL} proposed uncertainty-aware score distribution learning (USDL) to reduce the intrinsic ambiguity in the scores provided by experts.
Yu \etal \cite{CoRe} developed a contrastive regression (CoRe) framework, where they regress relative scores by referring to another video with shared attributes to learn, as well as a group-aware regression tree for assessment in a coarse-to-fine manner. 
Bai \etal \cite{aqa_tpt} proposed a temporal parsing transformer to decompose clip features into temporally ordered part features, which represent the temporal pattern for a specific action, to better capture the fine-grained intra-class variations, and assess action quality more accurately.

We evaluate our proposed method against SOTA AQA methods \cite{2020USDL,CoRe,aqa_tpt} on PFED5 and the recent SOTA pain estimation methods \cite{distanceordering2021,Szczapa_2022_trajec}
on the UNBC-McMaster Pain Dataset.
Furthermore, as we are using the RGB-based facial expression recognition approach Former-DFER\cite{formerDFER} for the extraction of RGB features in our model, we also evaluate it as a baseline by replacing its classification head with a regression head.


\section{PFED5 Facial Expression Dataset}
\label{sec:dataset}
\begin{figure}[t]
    \centering
    \includegraphics[width=0.90\linewidth]{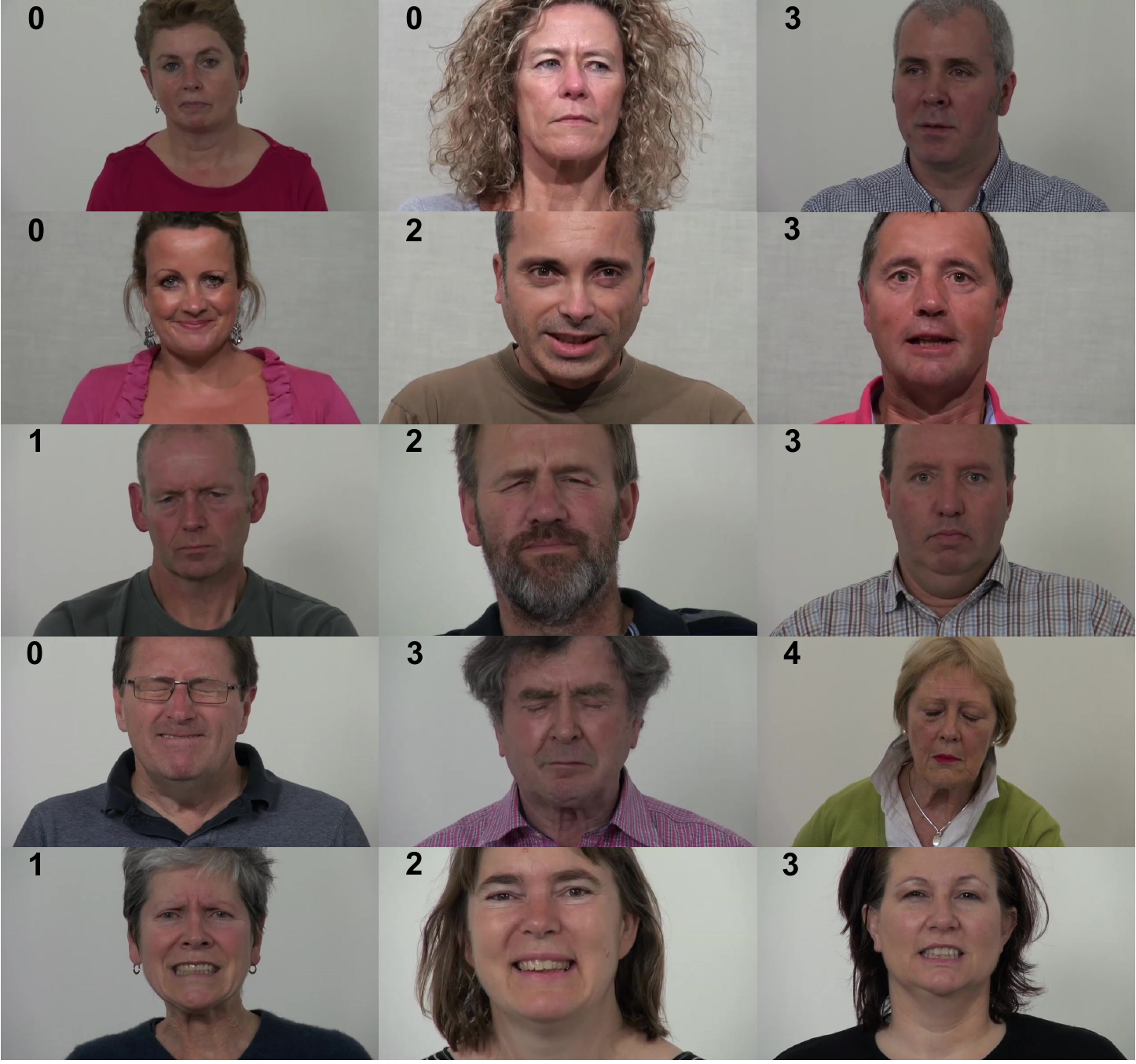}
    \caption{Example frames of different PD facial expressions in PFED5 at different severity levels. Rows top to bottom: sit at rest, smile, frown, squeeze eyes tightly, and clench teeth. {The number on the top left of each frame is the corresponding severity level: 0 (Normal), 1 (Slight), 2 (Mild), 3 (Moderate), and 4 (Severe). } }

    \label{fig:expression_example}
\end{figure}

{
The Parkinson's Disease Facial Expression Dataset (PFED5) contains 2811 video clips, covering 5 various facial expressions. 
The data was collected from 41
clinically-diagnosed PD patients {with 22 males and 19 females}, between 41 to 72 years old\footnote{All patients signed consent forms and full ethics approval for the release of the dataset is available.}. {The videos were captured at 25fps at a resolution of 1920×1080
(reduced to 854x480), using a SONY HXR-NX3 camera.} Each subject was required to perform 5 different actions: sit at rest, smile, frown, squeeze eyes tightly and clench teeth. {These actions were chosen as they reflect crucial factors influencing symptoms, including the frequency of blinking and the upper and lower facial movements.} The trained rater then assigned a score for each expression, based on the protocols of MDS-UPDRS~\cite{mds-updrs2008}, varying between 0 and 4 depending on the level of severity. Examples of the actions are shown in \Cref{fig:expression_example} and the number of videos available for each expression and at each severity level are presented in \Cref{tab:PD_statics}. We address the imbalance caused by the significantly fewer clips at severity level 4 through the choice of an appropriate loss function discussed later in the next section. 
}



\begin{table}
\centering
\resizebox{\columnwidth}{!}{
\begin{tabular}{lrrrrrrr} 
\toprule
\multirow{2}{*}{\begin{tabular}[c]{@{}l@{}}\textbf{Facial }\\\textbf{Expressions}\end{tabular}} & \multicolumn{5}{c}{\textbf{\#Videos of Severity Level}} & \multirow{2}{*}{\begin{tabular}[c]{@{}r@{}}\textbf{Min./Max./Avg. }\\\textbf{Seq. Len.}\end{tabular}} & \multirow{2}{*}{\begin{tabular}[c]{@{}r@{}}\textbf{\#Video }\\\textbf{Clips}\end{tabular}} \\ 
\cline{2-6}
 & \textbf{0} & \textbf{1} & \textbf{2} & \textbf{3} & \textbf{4} &  &  \\ 
\midrule
\multicolumn{8}{c}{\textit{\textbf{Training set}}} \\
Sit at rest & 104 & 126 & 78 & 90 & 11 & 10/84/35 & 409 \\
Smile & 87 & 129 & 126 & 61 & 12 & 27/219/80 & 415 \\
Frown & 86 & 147& 102 & 61 & 12 & 35/179/100 & 408 \\
Squeeze eyes & 106 & 153& 92 & 48 & 10 & 48/260/110 & 409 \\
Clench teeth & 53 & 110 & 171 & 68 & 4 & 50/210/113 & 406 \\ 
\midrule
\multicolumn{8}{c}{\textit{\textbf{Test set}}} \\
Sit at rest & 28 & 36 & 31 & 55 & 5 & 7/76/34 & 155 \\
Smile & 21 & 48 & 41 & 36 & 10 & 25/176/76 & 156 \\
Frown & 30 & 45 & ~25 & 31 & 21 & 40/228/100 & 152 \\
Squeeze eyes & 14 & 58 & 36 & 39 & 5 & 53/210/115 & 152 \\
Clench teeth & 18 & 39 & 64 & 23 & 5 & 63/247/118 & 149 \\ 
\midrule
\multicolumn{1}{c}{Total} & 547 & 891 & 766 & 512 & 95 & 7/260/88 & 2811\\
\bottomrule
\end{tabular}
}
\caption{{PFED5 dataset: the table shows 5 actions with the number of videos per action and at symptom severity levels 0-4, in both training and test sets. The severity levels are: 0 (Normal), 1 (Slight), 2 (Mild), 3 (Moderate), and 4 (Severe).}}

\label{tab:PD_statics}
\vspace*{-2mm}
\end{table}


\section{Approach}

\begin{figure*}[htb]
    \centering
    \includegraphics[width=0.92\textwidth]{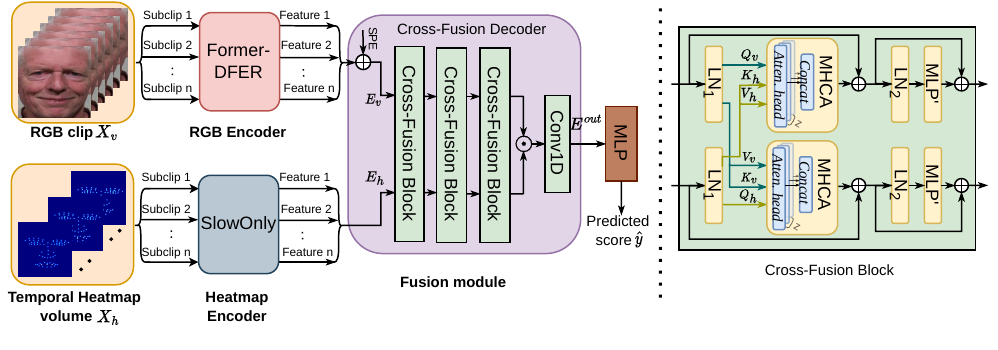}
    \caption{ 
    Overview of the proposed QAFE-Net. 
    Given a video sequence, we first pre-process it via the off-the-shelf face detector (SCRFD \cite{guo2021SCRFD}) to obtain facial landmark coordinates that are then used to generate the temporal heatmap volume. The RGB and landmark heatmap features are then encoded and a final 1D representation is obtained via the cross-fusion decoder, before passing through an MLP for predicting the clinical score. 
    LN: Layer Normalisation; MHCA: Multi-Head Cross-Attention; MLP: Multi-Layer Perceptron; SPE: Subclip Positional Embedding; $\oplus$ and $\odot$  indicate addition and  concatenation respectively. 
    } 
 
    \vspace{-3mm}
    \label{fig:pipeline}
\end{figure*}


We define a RGB video clip as $X_v \in \mathbb{R}^{T\times 3 \times H \times W}$, consisting of $T$ frames at height $H$ and width $W,$ and its corresponding temporal landmark heatmaps volume as $X_h  \in \mathbb{R}^{T\times 3 \times H \times W}$. 
Given $X_{v}$ and $X_{h}$ for a facial expression with symptom
level label $y$, our proposed QAFE-Net is formulated as a regression problem that predicts symptom level $\hat{y}$ as
\begin{equation}
    \hat{y } =\mathcal{\text{MLP}}( \mathcal{F}(\mathcal{P}(X_{v} ) , \mathcal{H}(X_{h} ) )) ~,
\end{equation}
where $\mathcal{P}$ and $\mathcal{H}$ denote the RGB and heatmap encoders, respectively, $\mathcal{F}$ is the fusion module, 
and MLP is the regression head.
The overall architecture of the proposed approach is illustrated in \Cref{fig:pipeline}. 

 {\bf RGB stream --} We adopt Former-DFER \cite{formerDFER} as our RGB encoder due to its SOTA performance, as well as the availability of its pre-trained weights on the large-scale facial expression dataset DFEW~\cite{DFEW}. 
It incorporates a convolutional spatial transformer and a temporal transformer to extract spatial and temporal features, respectively, {on a global scale}. As Former-DFER requires input sequences of 16 frames,
we prepare our input videos as follows: we randomly generate a clip of $T$ frames in a temporal order from each video, with the starting frame also chosen at random, with loop back to the beginning, if necessary. The clip is then split into $n$ subclips, such that $T=n \times t$ where $t=16$. 



\begin{figure}
  \centering
  \begin{subfigure}{0.75\linewidth}
    \includegraphics[width=0.9\linewidth]{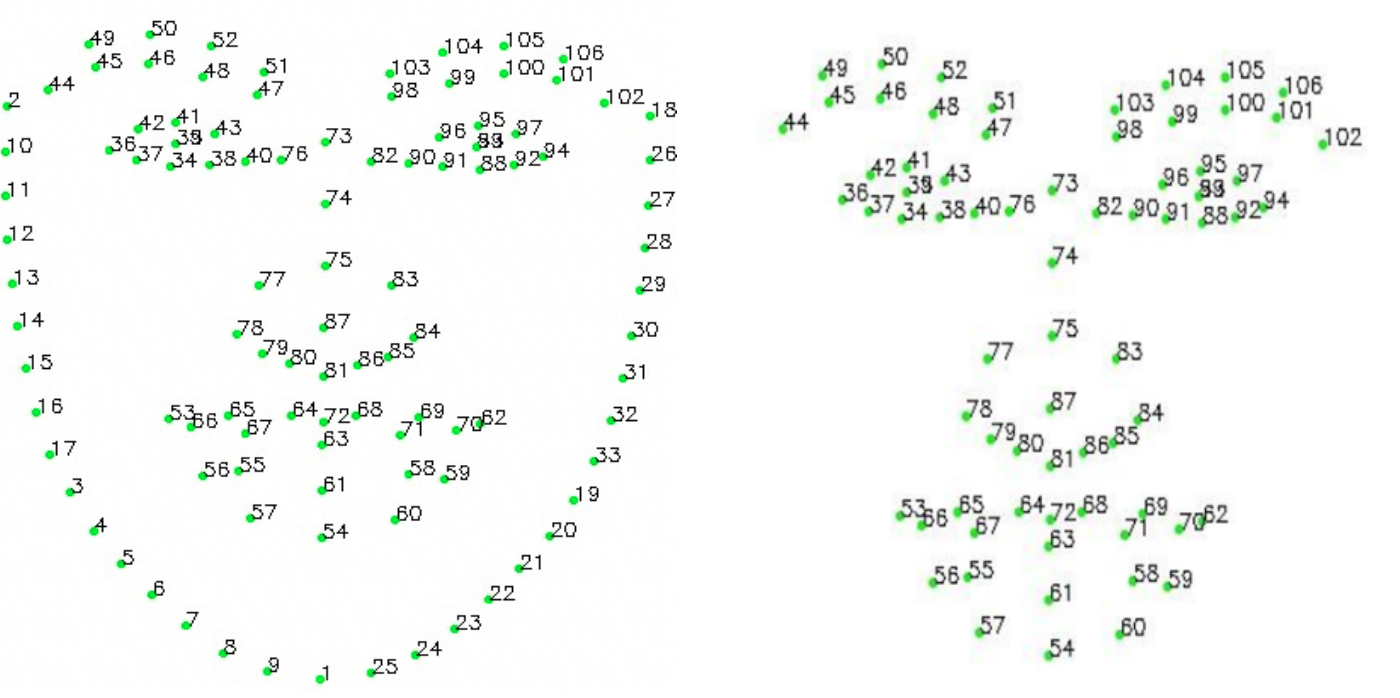}
    \caption{The 106 point landmarks (left) and the 73 points (right) mark-up for our method  where facial boundary landmarks are dropped.}
    \label{fig:short-a}
  \end{subfigure}
  \hfill
  \begin{subfigure}{0.75\linewidth}
    \includegraphics[width=0.9\linewidth]{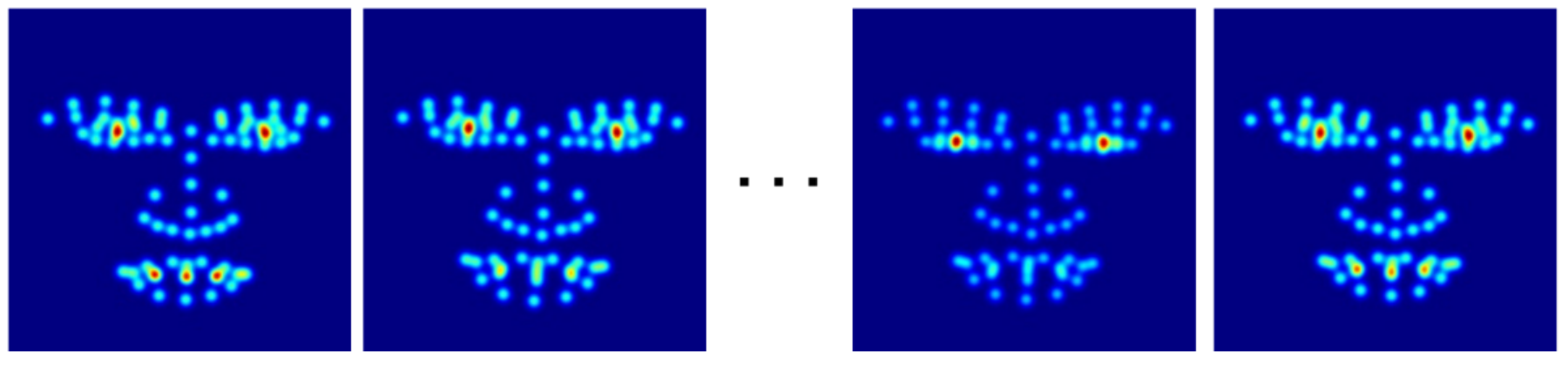}
    \caption{An example of a temporal heatmap volume from PFED5.}
    \label{fig:short-b}
  \end{subfigure}
  \hfill
  \begin{subfigure}{0.84\linewidth}
    \includegraphics[width=1.0\linewidth]{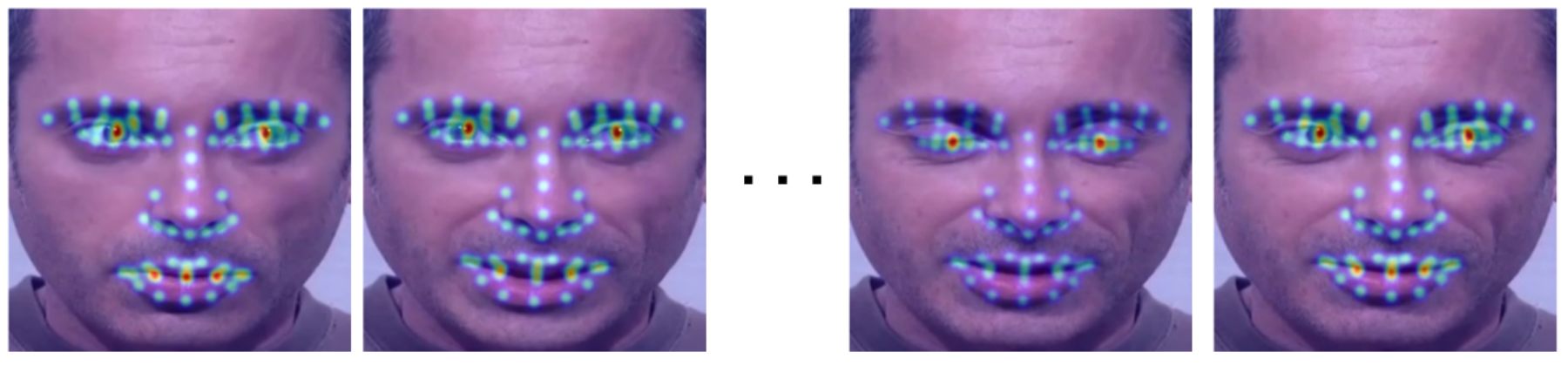}
    \caption{Visualisation of landmark heatmaps on facial regions.}
    \label{fig:short-c}
  \end{subfigure}
  \caption{Visualisation of Facial landmarks and heatmaps. }
  \label{fig:landmarks_visual}
  \vspace*{-3mm}
\end{figure}

 {\bf Landmark Heatmaps stream -- }
Inspired by the 3D heatmap volumes of skeleton joints proposed in \cite{PoseC3D_heatmap}, we propose temporal heatmap volumes at landmark positions to focus more locally on key facial movements associated with expressions. We select $M=73$ landmark points from the $106$ provided by \cite{guo2021SCRFD} (see \Cref{fig:short-a}). 


We pre-compute a Gaussian distribution of weights $G$ in an $11 \times 11$ kernel with a standard deviation of $\sigma$. This represents the decreasing weighted bounds of influence of a landmark around its position $(i,j)$. Then, we add the Gaussian at each landmark position while accumulating the weights into an accumulator frame ${\mathcal{A}}$ of size $H\times W$ to generate the initial heatmap, i.e.
\begin{equation}
  \mathcal{A}(i,j) =  \mathcal{A}(i,j) + G  ~~~~~ m=1..M  ~,
\end{equation}
{where $\mathcal{A}$ is initialised to zero. In our implementation, we set $\sigma=1$ and illustrate this choice through ablation.}
This accumulation of weights can lead to sharp peaks in $\mathcal{A}$, hence  we
{apply} a simple smoothing filter to it and
d{normalise} the weights to lie in the range $(0,1)$.
{These normalised weights are subsequently utilised to weight the RGB values, thereby generating the landmark heatmap for each frame.}
Our temporal heatmap volume $X_h$ is finally obtained by stacking all frames for a video clip along the temporal dimension (see example in \Cref{fig:short-b}).

To obtain heatmap feature embeddings, we use SlowOnly \cite{PoseC3D_heatmap} due to its SOTA performance on action recognition using skeleton joint heatmaps. These features are generated for $n$ subclips from $X_h$ in a corresponding fashion to the original video clip.


 {\bf Cross-Fusion Decoder --} 
{Cross-fusion modules were previously applied, for example,  in \cite{Zheng2022POSTERAP,finediving_tsa} to align spatial, temporal, and semantic correspondences between their (two) feature streams.} 
Zheng \etal \cite{Zheng2022POSTERAP} used a single cross-fusion block within their cross-fusion module in a pyramid structure. This design leveraged communication between their RGB  and landmark streams to enhance attention towards salient facial regions at multiple scales, which effectively mitigated issues related to inter-class similarity, intra-class discrepancy and scale sensitivity in FER when dealing with a single image. 
Our fusion module extends \cite{Zheng2022POSTERAP}'s to three cross-fusion blocks, as depicted in \Cref{fig:pipeline}, to prevent underfitting when processing data in video sequences and better address subtle facial motions that occur during an expression. We ablate this later in  \Cref{sec:ablation}.  

The first  stream into the fusion module is  the set of encoded RGB features $f_{v}=[f_{v}^{1},f_{v}^{2},...,f_{v}^{n}]$  added with subclip positional embeddings $E_{e}$  to give $E_{v} = f_{v} + E_{e}$, {where $ E_{e} =[E_{e}^{1},E_{e}^{1},...,E_{e}^{n}]$ and is used to ensure these features of $n$ subclips are in temporal order.}  The second stream is $E_{h}=[f_{h}^{1},f_{h}^{2},...,f_{h}^{n}]$, the set of landmark heatmap features from the $n$ subclips. The output of the fusion module $E^{out}$ is a concatenation of these two streams after they are processed through the three cross-fusion blocks and transformed into a 1D vector using a CONV1D block.

{The cross-fusion mechanism is achieved by the two Multi-Head Cross-Attention (MHCA) layers in each cross-fusion block, as shown in the right of  \Cref{fig:pipeline}. {Similar to \cite{Vaswani2017AttentionIA}, our use of multi-head attention empowers the model to concentrate on information from different representation subspaces simultaneously, improving the capability of the model to capture significant semantic information.} 
The input feature of each stream is initially subject to three linear transformations through the $\mathrm{LN_{1}} $ layer, resulting in three matrices {for each attention head}: Query $Q$, Key $K$, and Value $V$. 
{Within the cross-fusion process in the MHCA layer for $i^{th}$ attention head}, the query $Q_{vi}$ obtained from the RGB stream is compared with the key $K_{hi}$ from the landmark stream to find the relevance between them, and then the attention score is computed. Subsequently, the value $V_{hi}$ associated with $K_{hi}$ from the landmark stream is weighted and combined based on the attention score. {Then, all the feature maps from the $z$ attention heads are concatenated} to generate a new feature representation, i.e. 
\begin{equation}
    \mathrm{MHCA}_{v} = \mathrm{Concat}(\mathrm{head}_{v1},..,\mathrm{head}_{vz}) ~,
\end{equation}
\begin{equation}
    \mathrm{head}_{vi}=\mathrm {Softmax}(Q_{vi} \cdot K_{hi}^{T}/ \sqrt{d}) \cdot V_{hi} ~.
\end{equation}
Similarly, for the landmark stream, 
\begin{equation}
    \mathrm {MHCA}_{h}=\mathrm{Concat}(\mathrm{head}_{h1},..,\mathrm{head}_{hz}) ~,
\end{equation}
\begin{equation}
    \mathrm{head}_{hi}=\mathrm {Softmax}(Q_{hi}\cdot K_{vi}^{T}/ \sqrt{d})\cdot V_{vi} ~,
\end{equation}
\noindent where {$i=1,...,z$ and} $\frac{1}{\sqrt{d}}$  normalises to prevent vanishing gradients during back-propagation.

Finally, the output of fusion module $E^{out}$ is passed into an MLP to obtain a predicted score $\hat{y}$ of the facial expression quality.} {Here, we 
follow \cite{Vaswani2017AttentionIA} to employ $z = 8$ parallel attention heads.}

 {\bf Optimisation --}
To address the imbalanced regression in PFED5, we apply 
Batch-based Monte-Carlo (BMC)~\cite{balancedMSE_cvpr} as it has the advantage of estimating the training label distribution online and does not require additional prior information, making it easily applicable in real-world scenarios, 
\begin{equation}
    \mathcal{L}_{\mathrm {BMC}   }   =-log\frac{exp(-\left \| \hat{y}-y  \right \|_{2}^{2} /\tau  )}{ {\textstyle \sum_{y^{'}\in Y  }^{}} exp(-\left \| \hat{y}-y^{'}   \right \|_{2}^{2} /\tau  )} 
\label{eq:bmc}
\end{equation}
where $\tau=2\sigma^{2}_{noise}$ {is a temperature coefficient} and $Y$     
are the severity labels. In this work, $\sigma_{noise}=1$. We ablate our choice of BMC in Section~\ref{sec:ablation}.

\section{Experiments}  \label{sec:exp_setting}


{\bf Datasets --} Details regarding \textbf{PFED5} are reported in \Cref{sec:dataset}.  
{To balance the scores, 30 subjects were selected for training and 11 other subjects for inference}. 
\textbf{UNBC-McMaster} \cite{pain_dataset_2011} contains 200 videos from 25 subjects, ranging between 48-683 frames, 
with Visual Analog Scale (VAS) labels, which is a widely used self-reporting scale for pain estimation, ranging from 0 (no pain) to 10 (extreme pain).
Following  \cite{Szczapa_2022_trajec}, we evaluate using 5-fold cross-validation with 5 subjects per fold.




 \textbf{Evaluation Metric --}
As with past AQA works \cite{c3d_lstm_seven,2020USDL,CoRe, aqa_tpt}, we use Spearman’s Rank Correlation {$\rho$} for each expression in PFED5. 
For UNBC-McMaster Dataset, Mean Absolute Error (MAE) and Root Mean Square Error (RMSE) allow direct comparison with SOTA methods for pain estimation \cite{distanceordering2021, szczapa2020automatic,Szczapa_2022_trajec,xu2020AU}. 

 \textbf{Data Preparation --} We use SCRFD \cite{guo2021SCRFD} 
as the face detector to crop the facial region and detect landmarks for both datasets and save them as $256\times256$ images. 

 \textbf{Training Settings --} 
{Former-DFER \cite{formerDFER} is pretrained on DFEW \cite{DFEW} as the RGB encoder and SlowOnly \cite{PoseC3D_heatmap} is pretrained on NTU60-XSub \cite{Liu2019NTUR1} as the heatmap encoder.} The regression head is a three-layer MLP, with layers at $(512,256)$, $(256,128)$, and $(128,1)$.
We apply the SGD optimiser with momentum 0.9 and train  with a batch size of 4 for 100 epochs, initialising the learning rate as 0.001 and divide it by 10 every 40 epochs. 
 {During training, we randomly crop RGB clips to $224\times224$ and the temporal heatmap volumes to $56\times56$ and then apply horizontal flip to both inputs. }
We randomly sample 80 frames for each video in PFED5 and 144 frames in UNBC-McMaster, and partition them into 5 and 9 subclips respectively, with each subclip containing 16 frames. 
We introduce $5\times$ more iterations per epoch for data temporal augmentation, using random starting frames.
During inference, 80 frames as a clip are uniformly sampled from each video for PFED5. 
For UNBC-McMaster, we follow \cite{distanceordering2021} and randomly sample each video 10 times to generate 10 video clips, however here we use 144 frames as required by Former-DFER\cite{formerDFER}.
The final prediction is obtained by averaging the predictions of each clip. 


\subsection{Results on PFED5 Dataset}
\label{sec:pfed5_results}

We compare our QAFE-Net  on PFED5 with SOTA AQA methods, including C3D-LSTM\cite{learningscore_2016}, USDL\cite{2020USDL}, CoRe\cite{CoRe} 
and AQA-TPT\cite{aqa_tpt}, as well as SOTA dynamic FER methods\footnote{We simply replace their classification heads with a 3-layer MLP as a regression head.} Former-DFER \cite{formerDFER} and IAL-GCA \cite{GCA_IAL} (see \Cref{tab:sota_PD}). As the  source codes for SOTA methods on {\it pain estimation} are unavailable, we were unable to execute these methods on the PFED5 dataset for relative comparison. 

\begin{table*}[t]
\centering
\scriptsize
\begin{tabular}{llllcrrrrrr} 
\toprule
\textbf{Method} & \textbf{Publication} & \textbf{Backbone} & \textbf{Pretrained} & \textbf{Evaluator} & \begin{tabular}[c]{@{}c@{}}\textbf{Sit} \\\textbf{at rest}\end{tabular} & \textbf{Smile} & \textbf{Frown} & \begin{tabular}[c]{@{}c@{}}\textbf{Squeeze} \\\textbf{eyes}\end{tabular} & \begin{tabular}[c]{@{}c@{}}\textbf{Clench} \\\textbf{teeth}\end{tabular} & \begin{tabular}[c]{@{}c@{}}\textbf{Avg. }\\\textbf{Corr. }\end{tabular} \\ 
\hline
{C3D-LSTM\cite{c3d_lstm_seven}} & {CVPRW'17} & {C3D} & {UCF-101} & {LSTM} & {27.62} & {57.46} & {36.36} & {43.59} & {46.69} & {42.34} \\
{USDL\cite{2020USDL}} & {CVPR'20} & {I3D} & {K-400} & {USDL} & {40.08} & {55.90} & {46.36} & {44.45} & {52.17} & {47.79} \\
{CoRe\cite{CoRe}} & {ICCV'21} & {I3D} & {K-400} & {CoRe} & \underline{{50.23}} & {62.99} & {53.57} & {\underline{56.57}} & \underline{{65.05}} & {\underline{57.68}} \\
{AQA-TPT\cite{aqa_tpt}} & {ECCV'22} & {I3D} & {K-400} & {TPT} & {46.27} & {62.32} & {47.87} & {56.18} & \textbf{{67.51}} & {56.03} \\ 
\hline
{IAL-GCA\textsuperscript{$\dagger$}\cite{GCA_IAL}} & {AAAI'23} & {IAL-GCA} & {DFEW} & {MLP} & {30.83} & {\textbf{75.17}} & {43.25} & {38.23} & {59.02} & {49.30} \\
{Former-DFER\textsuperscript{$\dagger$}\cite{formerDFER}} & {ACMMM'20} & {Former-DFER} & {DFEW} & {MLP} & {43.77} & {67.32} & {48.89} & {47.88} & {50.91} & {51.75} \\
{QAFE-Net} & {-} & \begin{tabular}[c]{@{}l@{}}{Former-DFER+}\\{SlowOnly}\end{tabular} & \begin{tabular}[c]{@{}l@{}}{DFEW + }\\{ NTU60-XSub}\end{tabular} & {MLP} & \textbf{{75.57}} & {\underline{68.29}} & \textbf{{63.26}} & {\textbf{59.60}} & {60.01} & {\textbf{65.35}} \\
\bottomrule
\end{tabular}
\caption{Comparative Spearman's Rank Correlation results of QAFE-Net with SOTA AQA methods on PFED5. $\dagger$ indicates method originally proposed for dynamic FER, but with regression head added for our experiments. Best result is in bold, second-best is underlined.}
\label{tab:sota_PD}
\end{table*}

\begin{table*}[h]
\centering
\scriptsize
\begin{tabular}{lllrcrr} 
\toprule
\multirow{2}{*}{\textbf{Method}} & \multirow{2}{*}{\textbf{Description}} & \multirow{2}{*}{\textbf{Publication}} & \multirow{2}{*}{\begin{tabular}[c]{@{}c@{}}\textbf{\#Input}\\\textbf{ Frames}\textsuperscript{$\ddagger$}\end{tabular}} & \multirow{2}{*}{\textbf{Modality}} & \multicolumn{2}{c}{\textbf{VAS}} \\ 
\cline{6-7}
 &  &  &  &  & \textbf{MAE$\downarrow$} & \textbf{RMSE$\downarrow$} \\ 
\hline
 & \multicolumn{1}{l}{Human Rating (Observers' Pain Rating)\textsuperscript{$\xi$}} & & & Manual & 1.76 & 2.50 \\ 
\hline
Erekat~\etal \cite{erekat2020pain} & CNN-RNN & ICMIW'20 & 242 & RGB & 2.64 & - \\
Szczapa~\etal \cite{szczapa2020automatic}& Manifold trajectories & ICPR'20 & 60 & Landmarks\textsuperscript{$\diamond$} & 2.44 & 3.15 \\
Ting~\etal \cite{distanceordering2021} & R3D \& Distance Ordering & ICMLA'21 & 150 & RGB & 1.62 & \underline{2.10} \\
Szczapa~\etal \cite{Szczapa_2022_trajec} & Manifold trajectories \& split face regions  & TAFFC'22 & 121 & Landmarks\textsuperscript{$\diamond$} & \underline{1.59} & \textbf{1.98} \\ 
\hline
Kataoka \etal \cite{kataoka2020megascale} & R3D-18\textsuperscript{$\sharp$} & ArXiv'20 & 80 & RGB & 2.01 & 2.63 \\
Kataoka \etal\cite{kataoka2020megascale} & R3D-34\textsuperscript{$\sharp$} & ArXiv'20 & 80 & RGB & 2.07 & 2.61 \\
Zhao \etal\cite{formerDFER} & Former-DFER\textsuperscript{$\dagger$} & ACMMM'20 & 80 & RGB & 1.74 & 2.22 \\ \hline
Kataoka \etal\cite{kataoka2020megascale} & R3D-18\textsuperscript{$\sharp$} & ArXiv'20 & 144 & RGB & 1.92 & 2.42 \\
Kataoka \etal\cite{kataoka2020megascale} & R3D-34\textsuperscript{$\sharp$} & ArXiv'20 & 144 & RGB & 2.04 & 2.55 \\
Zhao \etal\cite{formerDFER}& Former-DFER\textsuperscript{$\dagger$} & ACMMM'20 & 144 & RGB & 1.67 & 2.21 \\ 
\hline
\multirow{3}{*}{QAFE-Net} & Former-DFER+SlowOnly (Cross-Fusion) & - & 144 & RGB+Landmarks\textsuperscript{$\star$} & 1.81 & 2.27 \\
& Former-DFER+SlowOnly (Concatenation) & - & 144 & RGB+Landmarks\textsuperscript{$\star$} & 1.98 & 2.54 \\
 & Former-DFER+SlowOnly (Summation) & - & 144 & RGB+Landmarks\textsuperscript{$\star$} & \textbf{1.57} & 2.11 \\

\bottomrule
\end{tabular}
\caption{Comparative MSE and RMSE results of QAFE-Net with SOTA Pain Estimation methods on  UNBC-McMaster.
$\dagger$  method originally proposed for dynamic FER, but  with regression head added for our experiments. $\diamond$ 66 facial landmarks from the original dataset. $\xi$ sequence-level score assigned by human raters  is used in \cite{pmlr-v116-xu20a} to estimate VAS. $\star$ the landmarks are regressed in the same way for PFED5. $\sharp$ model is pretrained on the combined dataset with Kinetics-700 and Moments in Time.  $\ddagger$ average values are used for experiments. Best result is in bold, second-best is underlined.}
\label{tab:sota_pain_vas}
\end{table*}

Both Former-DFER  and IAL-GCA  outperform previous non-contrastive regression methods C3D-LSTM and USDL, but score significantly lower than contrastive regression frameworks CoRe and AQA-TPT, as the latter better capture motions. 
After adding temporal landmark heatmap volumes as auxiliary learning, QAFE-Net achieves a substantial performance boost over RGB alone, at 65.35\%. {Other than on the `Clench teeth' and `Smile' actions}, QAFE-Net consistently outperforms other methods. {We conjecture that the model may erroneously flag anomalies when evaluating video clips that maintain an open mouth, particularly in these two actions.}

\begin{figure}[t]
\centering
    \includegraphics[width=0.48\linewidth]{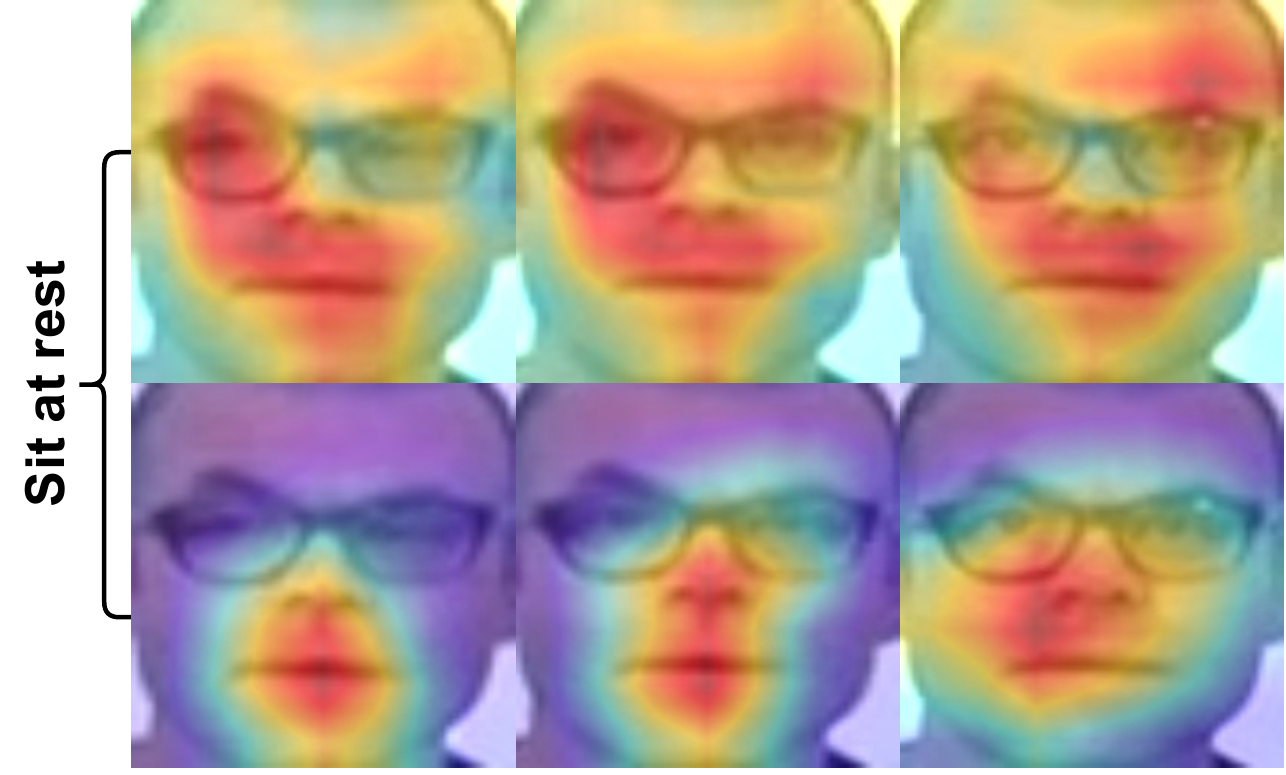}
    \includegraphics[width=0.48\linewidth]{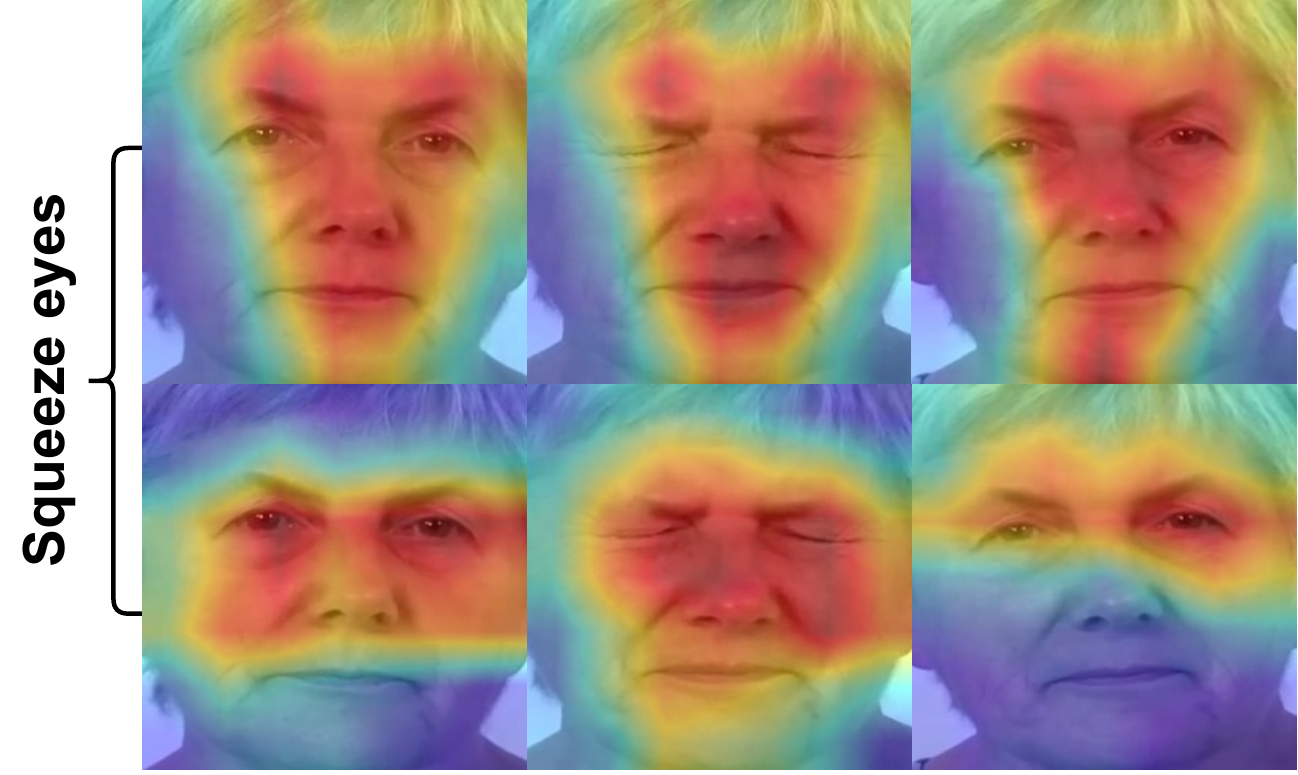} \\
    \includegraphics[width=0.48\linewidth]{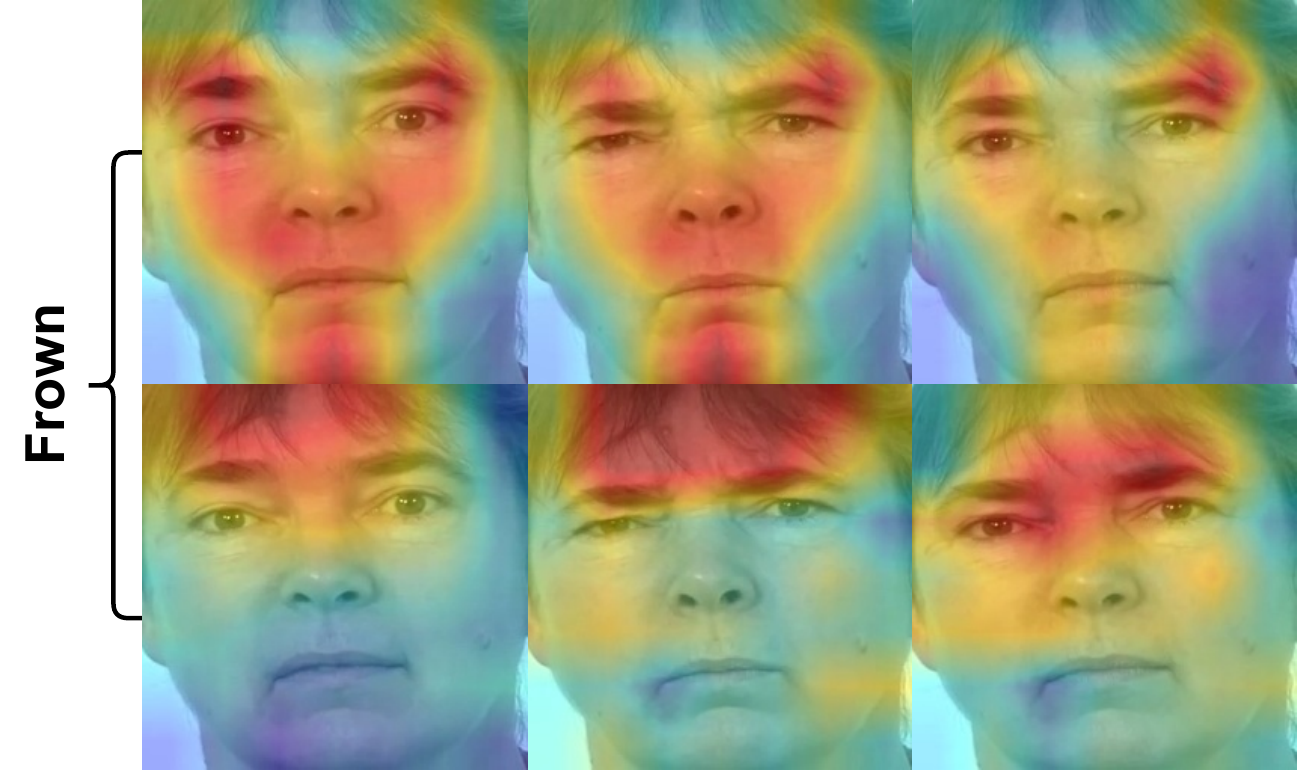} 
    \includegraphics[width=0.48\linewidth]{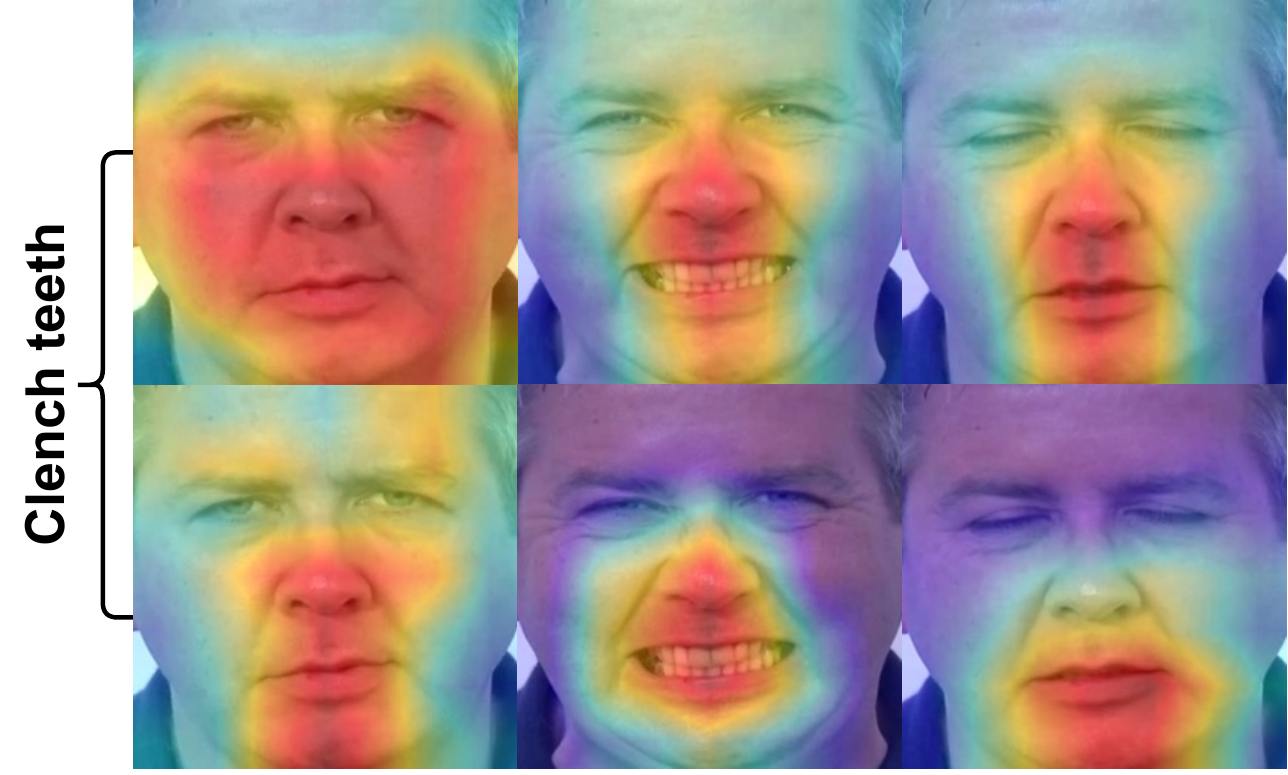} \\
    \includegraphics[width=0.48\linewidth]{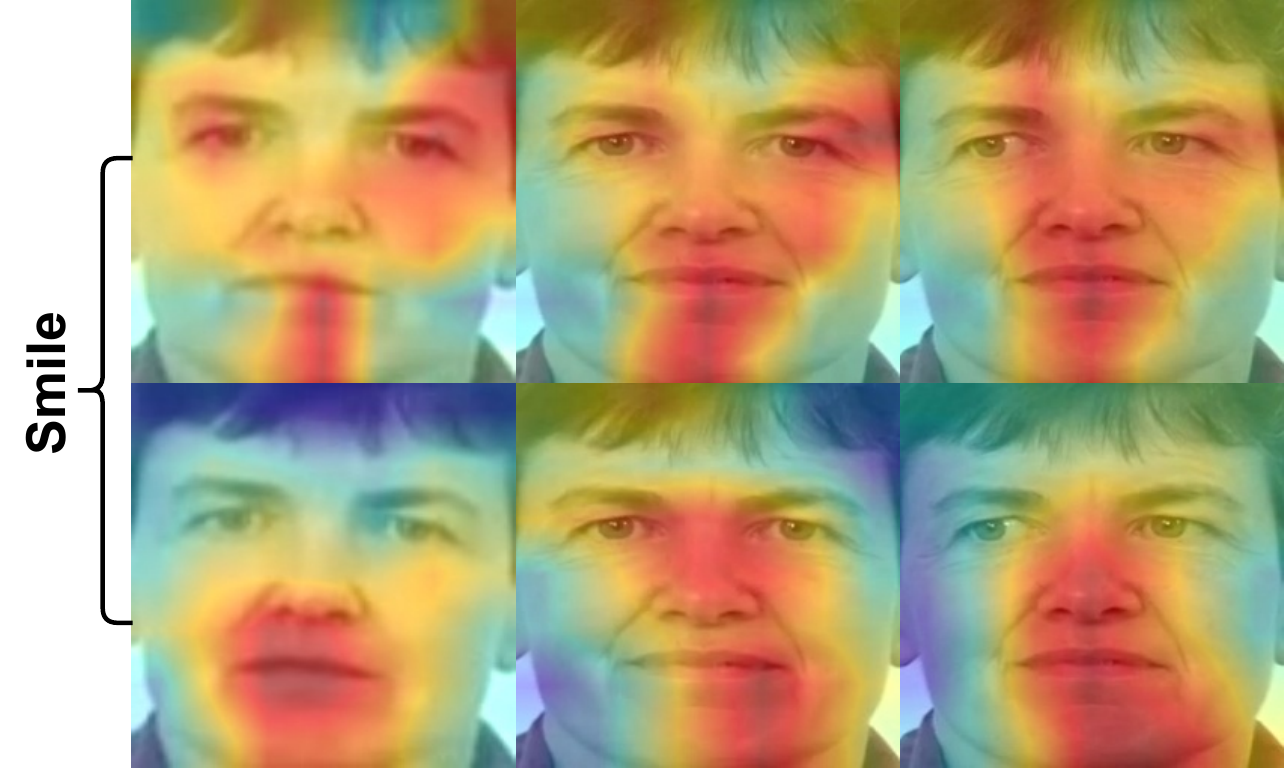}
    \includegraphics[width=0.48\linewidth]{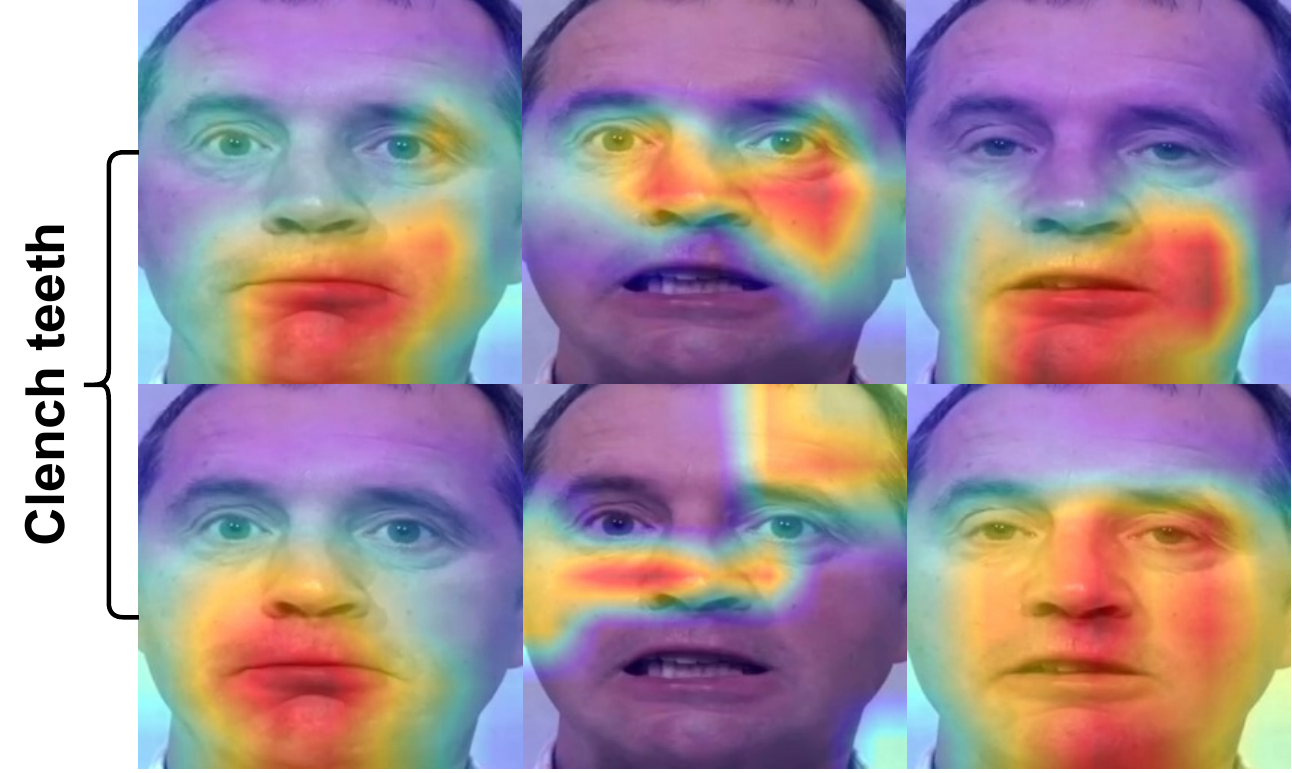}
\caption{Attention visualisation on PFED5. For each  action, top row corresponds to the RGB stream features, bottom row corresponds to landmark heatmap stream features. The lowest right visualisation depicts a failure case for the `Clench teeth' expression.} 
  \label{fig:attention_visual}
\end{figure}

{In \Cref{fig:attention_visual} we visualise example attention maps for the RGB and landmark streams in QAFE-Net. For each of our 5 expressions in PFED5, we can observe the attention learned by the landmark heatmap stream mainly emphasises on local regions (that contain landmarks), whereas the attention learned by the RGB stream is highlighted on a more global scale. A failure case for the {`Clench teeth' }expression is also shown in the lower-right of \Cref{fig:attention_visual}.}

\subsection{Results on UNBC-McMaster Dataset}
\label{sec:main_UNBC}

For this pain estimation dataset, we evaluate our proposed  QAFE-Net  against recent and SOTA pain estimation methods~\cite{erekat2020pain,szczapa2020automatic,distanceordering2021,Szczapa_2022_trajec}, as well as Former-DFER\cite{formerDFER}, and R3D-18\cite{kataoka2020megascale} and R3D-34\cite{kataoka2020megascale} baselines, again applied with a MLP as a regression head (see \Cref{tab:sota_pain_vas}).

Szczpa \etal \cite{Szczapa_2022_trajec}'s  landmarks based approach scores best at 1.98 RMSE, but the summation version of QAFE-Net obtains the lowest MAE at 1.57. We found that applying our cross-fusion module in QAFE-Net did not perform as well as simply summing our two feature embeddings. Through summation, more features from the non-landmark areas are available to the model, e.g. in {cheek areas which are significant regions for pain estimation~ \cite{pain_dataset_2011}}. While this helps minimise the error best in the summation version of QAFE-Net, it does not mean that landmark heatmaps are then redundant, since the RGB stream alone (e.g. our MLP-headed Former-DFER) does not perform better.

\subsection{Ablations}
\label{sec:ablation}
Next, we perform several ablations, all on the PFED5 dataset, examining the effects of landmark heatmaps, SlowOnly training, BMC loss, nature of the Gaussian shape, and other issues on QAFE-Net's performance. 

 \textbf{Effect of Temporal Heatmap Volumes --} Do we need temporal landmark heatmaps or would temporal landmarks alone be sufficient? This means no Gaussian weights for each landmark and its neighbourhood --  just  the landmark positions, as shown in the upper part of \Cref{fig:page1}.
As seen in \Cref{tab:ablation_landmarktype}, using temporal heatmap volumes does contribute significantly on most actions and on average ($\uparrow5.64\%$) to better capture subtle facial motions.

 {\textbf{Training SlowOnly --}  \Cref{tab:ablation_landmarktype} also shows that on average it is better to deploy a pretrained  SlowOnly on a large dataset rather than train it from scratch on a smaller target dataset, since the  former has learned more general patterns and features that potentially prove advantageous in scenarios involving limited data, such as PFED5, thereby serving as a mitigating factor against overfitting to a small dataset.}

\begin{table}[htb]
\centering
\scriptsize
\begin{tabular}{l|cccccc} 
\toprule
\multicolumn{1}{c|}{\textbf{Input}} & \begin{tabular}[c]{@{}c@{}}\textbf{Sit}\\\textbf{at rest}\end{tabular} & \textbf{Smile} & \textbf{Frown} & \begin{tabular}[c]{@{}c@{}}\textbf{Squeeze}\\\textbf{eyes}\end{tabular} & \begin{tabular}[c]{@{}c@{}}\textbf{Clench}\\\textbf{teeth}\end{tabular} & \begin{tabular}[c]{@{}c@{}}\textbf{Avg. }\\\textbf{Corr. }\end{tabular} \\ 
\hline
{Landmarks} & {49.81} & {\underline{67.77}} & {\underline{64.52}} & {\textbf{61.97}} & {\underline{54.46}} & {\underline{59.71}} \\
{Heatmaps\textsuperscript{$\star$}} & {\underline{52.59}} & {65.26} & {\textbf{65.43}} & {57.02} & {50.26} & {58.11} \\
{Heatmaps} & {\textbf{75.57}} & {\textbf{68.29}} & {63.26} & {\underline{59.60}} & {\textbf{60.01}} & {\textbf{65.35}} \\
\bottomrule
\end{tabular}
\caption{Effect of landmarks - Using facial landmarks alone (1st row) in QAFE-Net is not enough to capture the subtle motions also inscripted in the locality of the landmark points. Thus,  our landmark heatmaps (last row) fare better. 
$\star$ indicates that the SlowOnly used here was trained from scratch. All results are in \%. } 
\label{tab:ablation_landmarktype}
\end{table}

 \textbf{Effect of Cross-Fusion Decoder --} Although the outcome of the summation of the features was the best MAE result on the UNBC-McMaster dataset, in \Cref{tab:abla_fusion} we show that this does not hold true on PFED5 and the cross-fusion decoder is the best approach by a large margin. 

\begin{table}
\centering
\scriptsize
\begin{tabular}{l|cccccc} 
\toprule
\textbf{Fusion} & \begin{tabular}[c]{@{}c@{}}\textbf{Sit} \\\textbf{at rest}\end{tabular} & \textbf{Smile} & \textbf{Frown} & \begin{tabular}[c]{@{}c@{}}\textbf{Squeeze} \\\textbf{eyes}\end{tabular} & \begin{tabular}[c]{@{}c@{}}\textbf{Clench} \\\textbf{teeth}\end{tabular} & \begin{tabular}[c]{@{}c@{}}\textbf{Avg. }\\\textbf{Corr. }\end{tabular} \\ 
\hline
{Concatenation} & {69.93} & {65.49} & {45.98} & {\textbf{59.65}} & {\textbf{60.67}} & {\underline{60.34}} \\
{Summation} & {\underline{73.56}} & {\underline{66.52}} & {\underline{47.11}} & {54.28} & {58.99} & {60.09} \\
{Cross-Fusion} & \textbf{{75.57}} & \textbf{{68.29}} & \textbf{{63.26}} & {\underline{59.60}} & {\underline{60.01}} & {\textbf{65.35}} \\
\bottomrule
\end{tabular}
\caption{Comparison different fusion approaches in QAFE-Net on PFED5. As shown cross-fusion outperforms both summation and concatenation. All the results are in \%.}
\vspace{-4mm}
\label{tab:abla_fusion}
\end{table}

 \textbf{Effect of BMC Loss --} We also compare five different loss functions, including MSE, Focal-L1~\cite{imblancedregression_ICML}, Focal-MSE~\cite{imblancedregression_ICML}, GMM-based Analytical Integration (GAI)~\cite{balancedMSE_cvpr}, and BMC~\cite{balancedMSE_cvpr},  in the training stage of QAFE-Net. {In \Cref{tab:ablation_loss}, we observe that Focal-MSE, one of two modified versions of the Focal loss \cite{Lin2017FocalLF}, outperforms the MSE loss on PFED5. However, another version, Focal-L1, does not show the same level of performance. While GAI shows a marginal improvement over MSE, BMC emerges as the significantly better choice for handling class imbalance.}



\begin{table}[h]
\scriptsize
\centering
\begin{tabular}{l|cccccc} 
\toprule
\textbf{Loss} & \begin{tabular}[c]{@{}c@{}}\textbf{Sit}\\\textbf{at rest}\end{tabular} & \textbf{Smile} & \textbf{Frown} & \begin{tabular}[c]{@{}c@{}}\textbf{Squeeze}\\\textbf{ eyes}\end{tabular} & \begin{tabular}[c]{@{}c@{}}\textbf{Clench}\\\textbf{teeth}\end{tabular} & \begin{tabular}[c]{@{}c@{}}\textbf{Avg.}\\\textbf{ Corr.}\end{tabular} \\ 
\hline
{MSE} & {\underline{53.45}} & {67.13} & {59.69} & {57.19} & {51.27} & {57.75} \\
{Focal-L1\cite{imblancedregression_ICML}} & {50.98} & {66.19} & {61.84} & {56.40} & {51.61} & {57.40} \\
{Focal-MSE\cite{imblancedregression_ICML}} & {52.72} & {66.41} & {\underline{71.67}} & {57.76} & {51.86} & {60.08} \\
{GAI\cite{balancedMSE_cvpr}} & {52.56} & {\underline{68.66}} & {\textbf{73.55}} & {\textbf{60.00}} & {\underline{54.99}} & {\underline{61.95}} \\
{BMC\cite{balancedMSE_cvpr}} & \textbf{{75.57}} & \textbf{{68.29}} & {63.26} & {\underline{59.60}} & {\textbf{60.01}} & {\textbf{65.35}} \\
\bottomrule
\end{tabular}
\caption{Effect of using different loss functions in QAFE-Net on PFED5, with BMC~\cite{balancedMSE_cvpr}  performing best on average and GAI~\cite{balancedMSE_cvpr} recording second best. All the results are in \%.} 
\label{tab:ablation_loss}
\end{table}

 \textbf{Cross-Fusion Blocks --} 
\Cref{tab:ablation_CAblocks} summarises the effect of the number of cross-fusion blocks on model performance, with 1 to 5 blocks examined. QAFE-Net achieves its best average performance with 3 blocks, as well as the best performance on all expressions, bar one.  


\begin{table}[h]
\centering
\scriptsize
\begin{tabular}{ccccccc} 
\toprule
\textbf{Blocks} & \begin{tabular}[c]{@{}c@{}}\textbf{Sit }\\\textbf{at rest}\end{tabular} & \textbf{Smile} & \textbf{Frown} & \begin{tabular}[c]{@{}c@{}}\textbf{Squeeze}\\\textbf{ Eyes}\end{tabular} & \begin{tabular}[c]{@{}c@{}}\textbf{Clench}\\\textbf{ Teeth}\end{tabular} & \begin{tabular}[c]{@{}c@{}}\textbf{Avg.}\\\textbf{ Corr.}\end{tabular} \\ 
\hline
{1} & {71.11} & {53.66} & {\textbf{66.10}} & {57.47} & {57.29} & {61.13} \\
{2} & {74.03} & {64.48} & {\underline{65.27}} & {55.08} & {59.33} & {63.72} \\
{3} & \textbf{{75.57}} & \textbf{{68.29}} & {63.26} & {\textbf{59.60}} & {\textbf{60.01}} & {\textbf{65.35}} \\
{4} & {72.26} & {65.87} & {63.04} & {\underline{59.58}} & {59.32} & {64.11} \\
{5} & {\underline{74.84}} & {\underline{67.86}} & {62.56} & {58.86} & {\underline{59.50}} & {\underline{64.72}} \\
\bottomrule
\end{tabular}
\caption{Effect of different number of cross-fusion blocks in QAFE-Net's Cross-Fusion Decoder. Three blocks is the optimum number as shown by the results. All the results are in \%.}
\vspace{-4mm}
\label{tab:ablation_CAblocks}
\end{table}

 \textbf{Different feature outputs at $E_{out}$ --} Following the 3 cross-fusion blocks, we concatenate our features from the RGB and landmark streams prior to a CONV1D and send the resulting output vector $E_{out}$ to the MLP head. In \Cref{tab:ablation_output}, we examine different possible feature outputs from the fusion module:  with CONV1D (full method), without CONV1D, RGB stream only, and temporal heatmap stream only. We see that concatenating the outputs from both streams  improves the model performance over any single stream. 
 Further, the addition of the CONV1D layer helps capture the temporal dependencies present in the concatenated features, and  further boosts the model performance. 

\begin{table}
\centering
\scriptsize
\begin{tabular}{l|cccccc} 
\toprule
\begin{tabular}[c]{@{}l@{}}\textbf{Output}\\\textbf{Feature}\end{tabular} & \begin{tabular}[c]{@{}c@{}}\textbf{Sit }\\\textbf{at rest}\end{tabular} & \textbf{Smile} & \textbf{Frown} & \begin{tabular}[c]{@{}c@{}}\textbf{Squeeze}\\\textbf{ Eyes}\end{tabular} & \begin{tabular}[c]{@{}c@{}}\textbf{Clench}\\\textbf{ Teeth}\end{tabular} & \begin{tabular}[c]{@{}c@{}}\textbf{Avg.}\\\textbf{ Corr.}\end{tabular} \\ 
\hline
{RGB Stream} & {\underline{74.70}} & {\textbf{68.94}} & {58.70} & {\textbf{61.10}} & {58.38} & {64.36} \\
{Heatmap Stream} & {73.90} & {68.20} & {56.71} & {57.94} & {\underline{60.07}} & {63.38} \\
{Concat Stream} & {73.36} & {67.67} & {\underline{61.28}} & {59.25} & {\textbf{61.49}} & {\underline{64.61}} \\
\begin{tabular}[c]{@{}l@{}}{Concat Stream}\\{+ CONV1D\textsuperscript{$\dagger$}}\end{tabular} & {\textbf{75.57}} & {\underline{68.29}} & {\textbf{63.26}} & {\underline{59.60}} & {60.01} & {\textbf{65.35}} \\
\bottomrule
\end{tabular}
\caption{The performance comparison with using different output features from Cross-Fusion Decoder in QAFE-Net. All the results are in \%. Concat represents Concatenation. $\dagger$ represents the setting that can achieve best performance.}

\label{tab:ablation_output}
\end{table}

 \textbf{Classification on PFED5 --} We also apply QAFE-Net as a classification task for comparison with regression. Its best performance via focal loss is $57.03\%$, as shown in \Cref{tab:ablation_classify}, but this is  significantly lower than with regression at $65.35\%$ (see \Cref{tab:sota_PD}). Note that  Focal loss for classification improved its performance ($\uparrow 0.52\%$) compared to using Cross Entropy loss which is another indication that PFED5 has a class imbalance issue. {While the labels in PFED5 are discrete (same for UNBC-McMaster), the distance between categories carries significance. In contrast to classification tasks where the penalty for misclassification remains uniform across all classes, regression tasks impose penalties proportional to the distance from the correct label.}

\begin{table}
\scriptsize
\centering
\begin{tabular}{l|cccccc} 
\toprule
\textbf{Loss} & \begin{tabular}[c]{@{}c@{}}\textbf{Sit}\\\textbf{at rest}\end{tabular} & \textbf{Smile} & \textbf{Frown} & \begin{tabular}[c]{@{}c@{}}\textbf{Squeeze}\\\textbf{ Eyes}\end{tabular} & \begin{tabular}[c]{@{}c@{}}\textbf{Clench}\\\textbf{ Teeth}\end{tabular} & \begin{tabular}[c]{@{}c@{}}\textbf{Avg.}\\\textbf{ Corr.}\end{tabular} \\ 
\hline
{Cross Entropy} & {47.69} & {\textbf{65.03}} & {\textbf{64.55}} & {52.52} & {\textbf{52.74}} & {56.51} \\
{Focal\cite{Lin2017FocalLF}} & {\textbf{52.88}} & {59.99} & {61.84} & {\textbf{59.24}} & {51.18} & {\textbf{57.03}} \\
\bottomrule
\end{tabular}
\caption{Classification task with different losses on PFED5. The results show regression is a better choice for QAFE-Net.}
\vspace{-4mm}
\label{tab:ablation_classify}
\end{table}

 \textbf{Different $\sigma$ for generating landmark heatmaps --} 
\Cref{tab:ablation_gauss_scale} summarises the effect of $\sigma$, ranging from $1$ to $7$, in the Gaussian weights applied to each landmark for heatmap generation. Our method achieves its best performance at $\sigma=1$, and declines as $\sigma$ increases. This trend may be attributed to the increased density of weight overlap as $\sigma$ increases, thus enlarging the local areas that the model should  focus on. Thus, the wider the influence of the Gaussian weights is, the less ineffective are the landmark heatmaps. 

\begin{table}
\centering
\scriptsize
\begin{tabular}{ccccccc} 
\toprule
$\sigma$ & \begin{tabular}[c]{@{}c@{}}\textbf{Sit}\\\textbf{at rest}\end{tabular} & \textbf{Smile} & \textbf{Frown} & \begin{tabular}[c]{@{}c@{}}\textbf{Squeeze}\\\textbf{ Eyes}\end{tabular} & \begin{tabular}[c]{@{}c@{}}\textbf{Clench}\\\textbf{ Teeth}\end{tabular} & \begin{tabular}[c]{@{}c@{}}\textbf{Avg.}\\\textbf{ Corr.}\end{tabular} \\ 
\hline
N/A & {49.81} & {67.77} & {64.52} & {\textbf{61.97}} & {54.46} & {59.71} \\
$1$ & {\textbf{75.57}} & {68.29} & {63.26} & {\underline{59.60}} & {\textbf{60.01}} & {\textbf{65.35}} \\
$3$ & {52.76} & {\underline{69.01}} & {\underline{66.01}} & {58.32} & {\underline{56.59}} & {\underline{60.54}} \\
$5$ & {\underline{55.72}} & {68.60} & {\textbf{66.16}} & {58.04} & {52.92} & {60.29} \\
$7$ & {50.19} & {\textbf{70.67}} & {61.43} & {58.02} & {50.60} & {58.18} \\
\bottomrule
\end{tabular}
\caption{Effects of using different Gaussian spreads for weights in landmark heatmap generation. N/A denotes using only landmark positions (as shown in \Cref{fig:page1} and top row of \Cref{tab:ablation_landmarktype}) and so no weightings. Best results are obtained with $\sigma=1$.}
\vspace{-4mm}
\label{tab:ablation_gauss_scale}
\end{table}

\section{Conclusion}
 
{We introduced QAFE-Net, a method for facial expression quality assessment that explores global and local features, with the latter in particular facilitated by temporal heatmaps of influence instigated by facial landmarks. The spatio-temporal RGB and temporal landmark heatmap features are put through a cross-fusion decoder to examine and emphasise the relationship between them. Finally, the output is fed into a MLP head to predict the quality score. Our contributions include the introduction of temporal landmark heatmap representations and the development of the cross-fusion block for capturing both global and local expression characteristics from these two streams. Furthermore, we have introduced the fully annotated PFED5 dataset, a new facial expression benchmark for action quality assessment.  Our results on PFED5 and the UNBC-McMaster pain estimation benchmark underline the superiority of QAFE-Net, which outperforms SOTA methods for AQA on PFED5, while yielding the SOTA MAE result on UNBC-McMaster and comparable RMSE performance to that of leading pain estimation methods.
For future work, we aim to apply video synthesis  to generate additional artificial data for rare symptoms, addressing the class imbalance issue. Furthermore, audio cues from nurses are available in certain clinical situations, and this modality will be explored as an additional source of information.}


\section{Acknowledgements}
{The authors gratefully acknowledge the contribution of the Parkinson’s study participants and extend special appreciation to Tom Whone for his additional labeling efforts. The clinical trial from which the video data of the people with Parkinson’s was sourced was funded by Parkinson’s UK (Grant J-1102), with support from Cure Parkinson’s.}

{\small
\bibliographystyle{ieee_fullname}
\bibliography{egbib}
}

\end{document}